\documentclass[sigconf]{acmart}
\AtBeginDocument{%
  }

\setcopyright{none}              
\acmConference[]{}               
\acmYear{} \acmISBN{} \acmDOI{}  
\acmPrice{}                      
\usepackage{enumitem}

\begin{document}
\title{Laugh, Relate, Engage: Stylized Comment Generation for Short Videos}

\author{Xuan Ouyang}
\affiliation{%
  \institution{University of New South Wales}
  \city{Sydney}
  \country{Australia}
}
\email{xuan.ouyang@unsw.edu.au}

\author{Senan Wang}
\affiliation{%
  \institution{University of Sydney}
  \city{Sydney}
  \country{Australia}
}
\email{swan0472@uni.sydney.edu.au}

\author{Bozhou Wang}
\affiliation{%
  \institution{University of Sydney}
  \city{Sydney}
  \country{Australia}
}
\email{bwan0613@uni.sydney.edu.au}

\author{Siyuan Xiahou}
\affiliation{%
  \institution{The University of Hong Kong}
  \city{Hong Kong}
  \country{China}
}
\email{u3634276@connect.hku.hk}

\author{Jinrong Zhou}
\affiliation{%
  \institution{University of Southern California}
  \city{Los Angeles}
  \country{USA}
}
\email{jinrongz@usc.edu}

\author{Yuekang Li}
\affiliation{%
  \institution{University of New South Wales}
  \city{Sydney}
  \country{Australia}
}
\email{yuekang.li@unsw.edu.au}









\begin{abstract}
Short-video platforms have become a core medium of the modern Internet, where efficient information delivery and strong interactivity reshape user engagement and cultural diffusion. Comments, as a primary form of user expression, play a critical role in fostering community activity and content re-creation. However, generating comments that are both platform-compliant and capable of exhibiting stylistic diversity with contextual awareness remains a challenging task. We present \textbf{LOLGORITHM}, a modular multi-agent system (MAS) for controllable short-video comment generation. The system integrates video segmentation, contextual and affective analysis, and style-aware prompt construction, supporting six distinct comment styles: puns (homophones), rhyming, meme application, sarcasm (irony), plain humor, and content extraction. Powered by a multimodal large language model (MLLM), \textbf{LOLGORITHM} directly processes video inputs and achieves fine-grained style control through explicit prompt markers and few-shot examples. We construct a bilingual dataset from Douyin (Chinese) and YouTube (English) via official APIs, covering five popular video genres: comedy skits, daily life jokes, funny animal videos, humorous commentary, and talk shows. Evaluation combines automated metrics—originality, relevance, and style conformity—with a large-scale human preference study involving 40 videos and 105 participants. Results show that \textbf{LOLGORITHM} significantly outperforms baseline models, achieving preference rates of over 90\% on Douyin and 87.55\% on YouTube. This work introduces a scalable and culturally adaptive framework for stylized comment generation on short-video platforms, providing a promising path to enhance user engagement.
\end{abstract}




\maketitle

\section{Introduction}


In recent years, short video platforms have become an integral part of the Internet, and short videos have emerged as a crucial channel for multimedia information dissemination. As an essential component of short videos, comments are not only an important outlet for users to express their views, but also play a key role in content propagation, community interaction, and algorithmic feedback. With the development of generative artificial intelligence, comments are no longer merely the endpoint of user expression, but have become the starting point for driving content evolution and platform ecosystem circulation. Short video comments are transforming from auxiliary information to core driving forces in the Web ecosystem: they not only carry user viewpoints but also participate in content re-creation, community interaction, and algorithmic feedback, becoming key nodes for platforms to understand users, optimize recommendations, and stimulate dissemination. With the intervention of multimodal large language models, comment sections are reshaping information flow patterns, community vitality, and semantic structures, enabling the Web to transition from static display to dynamic participation and contextual evolution.


Although current multimodal large language models possess the capability to handle complex tasks, they remain inadequate in short video comment generation. As the role of comments in enhancing platform engagement becomes increasingly prominent, this area has yet to be fully explored. Short video comments exhibit rich and diverse stylistic forms—ranging from memes and irony to rhyming and puns—posing severe challenges to generation systems in terms of style controllability and sociolinguistic adaptability. How to generate comments that are contextually appropriate, stylistically diverse, and aligned with platform-specific community norms has become a critical gap in current multimodal content generation capabilities.


Current research related to short video comment generation primarily focuses on two directions: video summarization generation and live-streaming comment generation. Video summarization generation emphasizes semantic compression while neglecting the stylistic expressiveness required for comments, resulting in generated content that lacks appeal and interactivity. Live-streaming comment generation typically relies on dynamic user input and continuous context, making it difficult to transfer to structurally compact, information-dense standalone short video scenarios. Its generation objectives also lean more toward guiding live-streaming interactions, lacking the guiding capability required for short video comments and failing to drive content evolution and platform ecosystem circulation. Meanwhile, existing methods generally lack support for multiple controllable styles and struggle to generate comment content that is both stylistically diverse and aligned with platform-specific culture and linguistic nuances, showing significant deficiencies in meeting authentic community contexts.


We propose LOLGORITHM, a novel extensible modular multi-agent architecture specifically designed for short video comment generation. This framework can generate six controllable comment styles: puns (homophones), rhyming, meme application, sarcasm (irony), plain humor, and content extraction, with style guidance through explicit prompt markers and few-shot examples. The system uses multimodal large language models (MLLMs) as the generation backbone to achieve multimodal input fusion and conditional text generation. To accomplish cross-platform tasks, we have constructed two short video comment datasets from different platforms to enhance comment contextual fit, stylistic diversity, and community adaptability.


The comment generation component of LOLGORITHM consists of three main modules: video content extraction, video content classification, and comment generation, capable of processing multimodal video content to generate platform-appropriate comments. To capture stylistic variations across communities, we have compiled a bilingual dataset from Douyin (Chinese) and YouTube (English), covering five high-engagement video types: comedy skits, daily life jokes, funny animal videos, humorous commentary, and talk shows. Each sample has been pre-screened to ensure visual clarity and conversational relevance. The agents leverage platform-specific style priors and adapt to community norms during the generation process.


We employ a dual approach combining automated metrics and human preferences for comprehensive evaluation. Our automated scoring framework assesses comments across three dimensions: originality, relevance, and style conformity. Results demonstrate that LOLGORITHM achieves leading scores on both Douyin and YouTube. Meanwhile, human evaluation involving 40 videos and 105 respondents shows strong user preference for LOLGORITHM-generated comments, with selection rates exceeding 90\% on Douyin and reaching 87.55\% on YouTube.


The main contributions of this work are as follows:

\begin{itemize}[leftmargin=*]
    \item We introduce a novel modular multi-agent architecture for controllable short video comment generation. To our knowledge, this is the first system specifically designed for short video comment generation.
    
    \item We provide a comprehensive bilingual dataset covering multiple video types and comment styles, establishing a solid foundation for training and evaluation.
    
    \item Through extensive empirical analysis, we demonstrate the effectiveness of style control, modular prompting, and multimodal context modeling in generating engaging, platform-aligned comments that are consistent with real-world user engagement patterns.
\end{itemize}

To facilitate reproducible research, we provide code and raw experimental data as supplementary materials. An anonymous GitHub repository containing all relevant resources is publicly available at \href{https:https://anonymous.4open.science/r/for-review-505F}{https://anonymous.4open.science/r/for-review-505F}.
\section{Related Work}

\subsection{Video Summarization}
Apostolidis et al.\cite{Apostolidis2021VideoSU} surveyed over 40 deep learning-based video summarization methods, encompassing supervised, weakly supervised, unsupervised, and multimodal approaches. They proposed a unified classification framework, analyzing strategies in terms of extraction efficiency, contextual modeling, and scene adaptability. Otani et al.\cite{Otani2022VideoSO} defined video summarization as extracting key information from long videos to improve viewing efficiency across domains, introducing a taxonomy based on Domains, Purposes, and Formats to guide standardization. Tonge et al.\cite{Tonge2022ANA} developed a static summarization method using keyframe extraction and storyboard generation, improving narrative coherence and user engagement. Yao et al.\cite{Yao2022MultiLevelSN} proposed a multi-level spatiotemporal framework that jointly models frames, fragments, and shots, showing strong industrial applicability. Addressing data scarcity, Hua et al.~\cite{Hua2024V2XumLLMCV} introduced Instruct-V2Xum, a dataset of 30{,}000 YouTube videos (40–940s) with frame-referenced summaries and a 16.39\% compression ratio, offering high semantic alignment for multimodal research.

\subsection{Live-Streaming Danmaku Generation}
Research on live-streaming danmaku generation has progressed from corpus construction to multimodal and personalized comment modeling. Ma et al.\cite{Ma2018LiveBotGL} pioneered the field with a Bilibili corpus and the LiveBot framework. Wang et al.\cite{Wang2020VideoICAV} introduced the denser VideoIC dataset and MML-CG, a multimodal multitask model surpassing LiveBot in accuracy and diversity. Sun et al.\cite{Sun2023ViCoEV} incorporated like'' counts in ViCo-20k to reflect human preferences. Fang et al.~\cite{Fang2020Video2CommonsenseGC} broadened semantic scope via commonsense-grounded comment generation. Lalanne et al.~\cite{Lalanne2023LiveChatVC} proposed a Triple Transformer Encoder using visual, audio, and textual contexts, alongside the English LiveChat dataset. Luo et al.~\cite{Luo2024EngagingLV} introduced socially engaging comment generation with a multimodal dataset annotated with like'' signals. Wu et al.\cite{Wu2024UnderstandingHP} integrated user preference modeling and video segmentation to enhance alignment and anthropomorphism. Lin et al.\cite{Lin2024PersonalizedVC} proposed PerVidCom for personalized video comment generation, emphasizing style migration and semantic alignment. Gao et al.\cite{Gao2023LiveChatAL} focused on persona-driven interaction using a large-scale dialogue dataset with host speech, user comments, and profiles.

\subsection{Research Gap}
Despite advances in video summarization and live-streaming danmaku generation, directly applying these techniques to short video comment generation remains challenging. First, summarization methods focus on compressing core content, often overlooking user opinions, emotional cues, and social context---key elements that make comments engaging. Existing video summarization methods lack emotional naturalness and stylistic nuance, resulting in generated content that is overly monotonous, exhibits low style consistency, and lacks the interactivity that comments should possess. Although some research in live-streaming danmaku generation has explored personalization, current models heavily rely on historical user behavior within single-user, single-stream contexts, which limits generalization capability across users and content types, especially in cross-platform scenarios. Moreover, such frameworks exhibit weak robustness: when faced with short videos containing excessively large and complex amounts of information, they fail to generate comments and can only repeatedly output fixed replies from predefined templates. Furthermore, platform-specific designs hinder adaptability to varying content styles, linguistic norms, and interaction patterns across short video platforms. Finally, existing methods are mostly limited to a single language, lacking frameworks that can be used across multiple languages or on different language platforms. Additionally, there are no suitable datasets for short video comments, necessitating the creation of datasets required for training.
\section{Methodology}

\begin{figure*}[t]
    \centering
    \includegraphics[width=\linewidth]{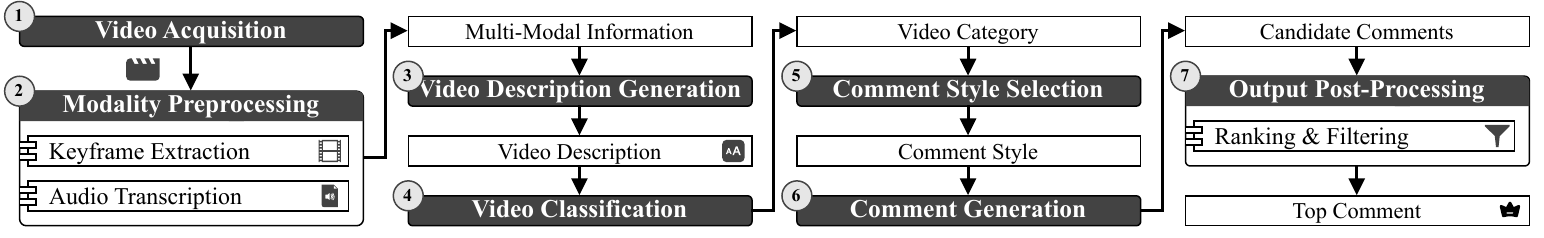}
    \caption{Workflow of the LOLGORITHM framework.}
    \Description{A flowchart showing the three-stage pipeline: data preprocessing, dataset construction, and comment generation, with arrows indicating data flow between modules.}
    \label{fig:framework}
\end{figure*}

LOLGORITHM is a novel, scalable, modular, multi-platform, and multilingual short video comment generation agent, as well as a novel framework. Currently, LOLGORITHM can primarily generate Chinese and English comments for Douyin and YouTube platforms. This section provides a detailed description of LOLGORITHM, introducing it from three parts: the data preprocessing module, the dataset construction module, and the comment generation module. Figure~\ref{fig:framework} presents the complete workflow of the LOLGORITHM framework. The system consists of three main components: the data preprocessing module, the dataset construction module, and the comment generation module. Each component is described in detail in the following subsections.

\subsection{Data Preprocessing Module}

In the entire comment generation system, the data preprocessing module plays a crucial role. It not only provides structured semantic input for subsequent comment generation but also serves a foundational role in the dataset construction process. This module consists of three sub-modules: the information download module, the video processing module, and the video description module. These three components work collaboratively to complete the modal transformation from raw video to text description, laying a solid foundation for the system's intelligent generation.

First, the entire workflow begins with creating a JSON file containing the target video URLs. This file serves as the entry point for data preprocessing, recording basic information about the videos to be processed. Subsequently, this JSON file is input into the information download module. This module automatically completes the download of video content by calling the official APIs provided by Douyin and YouTube. Video files are typically saved to a local specified directory in MP4 format, while video descriptions are written into the original JSON file and stored as video metadata. This step not only ensures data integrity but also provides the necessary semantic background for subsequent processing.

Next, the downloaded videos are input into the video processing module. The core task of this module is to transform video content into analyzable textual information. The processing workflow is divided into multiple stages, with the first step being highlight detection. The system employs a dual detection mechanism of audio fluctuation and video light source variation to identify exciting segments in the video. Let $A(t)$ represent the audio amplitude at time $t$ and $L(t)$ represent the light intensity. The highlight score $H(t)$ at time $t$ is computed as:

\begin{equation}
H(t) = \omega_a \cdot \left|\frac{dA(t)}{dt}\right| + \omega_l \cdot \left|\frac{dL(t)}{dt}\right|
\end{equation}

where $\omega_a$ and $\omega_l$ are weights for audio and light variations respectively. A segment is identified as a highlight if $H(t) > \theta_h$, where $\theta_h$ is a predefined threshold.

After highlight detection is completed, the system performs frame extraction on the video. The frame extraction strategy is differentially designed based on the importance of video segments. The frame extraction rate $r(t)$ at time $t$ is defined as:

\begin{equation}
r(t) = \begin{cases}
10 \text{ fps} & \text{if } H(t) > \theta_h \text{ (highlight segment)} \\
0.5 \text{ fps} & \text{otherwise (normal segment)}
\end{cases}
\end{equation}

These extracted images are saved to a specified location and recorded in the JSON file along with audio transcription information. Audio transcription uses automatic speech recognition technology to convert speech content in the video into text and associates it with corresponding image sequence numbers, forming complete timeline semantic data.

Complete the above processing, the system inputs the JSON file containing the video description, audio transcription text, and extracted frame image paths into the video description module. This module calls a multimodal large language model to comprehensively analyze text information and image content. Let $\mathcal{T} = \{t_1, t_2, \ldots, t_n\}$ represent the set of transcribed text segments and $\mathcal{F} = \{f_1, f_2, \ldots, f_m\}$ represent the set of extracted frames. The video description $D$ is generated by:

\begin{equation}
D = \text{LLM}_{\text{multimodal}}(\mathcal{T}, \mathcal{F}, \text{meta})
\end{equation}

where $\text{meta}$ represents the video metadata including title, description, and duration. Through this process, the system achieves modal transformation from video to text, providing high-quality semantic input for subsequent video classification and comment generation modules.

\subsection{Dataset Construction Module}

In the current research field of short video content generation, comment generation, as an important form of user interaction, has gradually received widespread attention from academia and industry. However, existing research generally faces a critical bottleneck—the lack of high-quality, structured short video comment datasets, especially cross-platform Chinese-English bilingual datasets. This deficiency not only limits the training and evaluation of multilingual models but also hinders the generalization capability and practicality of comment generation systems on a global scale.

To address this problem, we constructed a Chinese-English bilingual dataset specifically for cross-platform short video comment research. The Chinese comment portion of this dataset comes from Douyin, while the English comment portion is collected from YouTube. These two platforms are respectively the short video platforms with the largest number of users and highest activity levels in the Chinese and English language systems, possessing strong representativeness and coverage. Choosing these two platforms as data sources not only reflects user expression habits under different linguistic and cultural backgrounds but also provides a solid corpus foundation for training multilingual comment generation models.

In terms of data collection, we selected a total of 200 short videos as samples, with Douyin and YouTube each accounting for 100 videos. Let $N_{\text{total}} = 200$, $N_{\text{Douyin}} = 100$, and $N_{\text{YouTube}} = 100$. To ensure content diversity and comment interactivity, we performed balanced sampling from five video types with high user engagement:

\begin{enumerate}[leftmargin=*]
    \item \textbf{Talk Show}: Videos featuring conversational content with hosts or speakers addressing audiences.
    \item \textbf{Humorous Commentary}: Videos containing comedic narration or commentary on various topics.
    \item \textbf{Funny Animal}: Videos showcasing amusing animal behaviors and interactions.
    \item \textbf{Daily Life Jokes}: Videos depicting humorous situations from everyday life.
    \item \textbf{Comedy Skits}: Short scripted comedic performances and sketches.
\end{enumerate}

Each type has 20 videos selected on each platform. The distribution can be expressed as:

\begin{equation}
N_{p,c} = \frac{N_p}{|\mathcal{C}|} = \frac{100}{5} = 20
\end{equation}

where $N_{p,c}$ is the number of videos for platform $p$ and category $c$, and $|\mathcal{C}| = 5$ is the number of categories, ensuring balance and representativeness in the dataset's content distribution.

The semantic description of video content refers to the data preprocessing module described above. Specifically, we obtain videos and their descriptions through the information download module, perform highlight detection, frame extraction, and audio transcription through the video processing module, and finally call a multimodal large language model in the video description module to generate structured video content descriptions. These descriptions not only provide semantic input for comment generation but also constitute training data for video classification models. Ultimately, we obtained a semantic dataset consisting of 200 video content descriptions, laying the foundation for subsequent classification and generation tasks.

Comments are obtained using official APIs provided by the platforms, calling Douyin and YouTube's comment data services respectively to automatically download the top five comments with the highest number of likes for each video. For video $v_i$, let $\mathcal{L}(v_i) = \{l_1, l_2, \ldots, l_k\}$ be the set of like counts for all comments. We select the top-$K$ comments where $K=5$:

\begin{equation}
\mathcal{C}_{\text{selected}}(v_i) = \{c_j \mid l_j \in \text{Top-K}(\mathcal{L}(v_i)), j = 1, \ldots, K\}
\end{equation}

This strategy ensures the representativeness and interactivity of collected comments while also avoiding legal risks and ensuring the data collection process is legal and compliant. The total number of comments in the dataset is:

\begin{equation}
N_{\text{comments}} = N_{\text{total}} \times K = 200 \times 5 = 1000
\end{equation}

The downloaded comments are saved together with the corresponding video content descriptions in a new JSON file, forming a structured comment dataset.

To further enhance the research value of the dataset, we conducted manual annotation of all 1,000 comments, analyzing their generation methods and language styles. The annotation process was jointly completed by multiple researchers with experience in linguistics and short video content analysis, ensuring the accuracy and consistency of annotation results. We categorized the comment generation methods into six major types:

\begin{enumerate}[leftmargin=*]
    \item \textbf{Puns (Homophones)}: Creating humorous effects using homophones or near-homophones, commonly seen in Chinese comments.
    \item \textbf{Rhyming}: Enhancing language rhythm through rhyme, improving the readability and entertainment value of comments.
    \item \textbf{Meme Application}: Citing internet popular phrases or memes, enhancing the cultural resonance and dissemination power of comments.
    \item \textbf{Sarcasm (Irony)}: Using satire or irony to express viewpoints, commonly used in teasing or critical comments.
    \item \textbf{Plain Humor}: Directly expressing humorous content without relying on linguistic techniques or cultural background.
    \item \textbf{Content Extraction}: Directly extracting information from video content for commenting, reflecting the semantic fit of the comment.
\end{enumerate}

Let $\mathcal{S} = \{s_1, s_2, \ldots, s_6\}$ denote the set of six style categories. Each comment $c_i$ is assigned a style label:

\begin{equation}
\text{label}(c_i) \in \mathcal{S}
\end{equation}

Through this systematic annotation process, we not only obtained language style labels for comments but also provided supervision signals for subsequent comment generation models. This short video comment dataset thus possesses characteristics of multilingualism, multiple types, and multiple styles, capable of supporting training and evaluation of various comment generation tasks, including style transfer, semantic fit optimization, and cross-language generation.

The Chinese-English bilingual short video comment dataset we constructed fills the gap in existing research, possessing high representativeness, structure, and scalability. It not only provides rich training corpora for comment generation models but also provides a solid foundation for multilingual content understanding and generation research. In the future, we plan to further expand the scale and types of the dataset, introducing more languages and platforms to promote diversified development and global application of short video comment generation technology.

\subsection{Comment Generation Module}

The comment generation module of LOLGORITHM is the core component of the entire system, undertaking the critical task of transforming video content into high-quality, style-matched original comments. This module consists of two main sub-modules: the video classification module and the comment generation module. Its workflow is tightly connected, relying on prior data preprocessing and semantic matching mechanisms to ensure that generated comments are both relevant to video content and possess high style consistency and originality.

First, the system performs modal conversion on the target video through the data preprocessing module, i.e., transforming video content into structured text descriptions. This process typically relies on the collaborative work of multimodal technologies such as video subtitles, speech recognition, and image recognition, extracting core semantic information from the video. This information includes not only explicit elements such as characters, scenes, and actions in the video but also implicit features such as emotions, tone, and rhythm, providing a rich semantic foundation for subsequent classification and comment generation.

Next, the video content description is input into the video classification module. The main task of this module is to determine which category the target video belongs to through semantic matching. To achieve this goal, the system pre-constructs a dataset containing a large number of video content descriptions, each associated with a clear video category. The classification module calculates semantic similarity between the target video's text description and all samples in the dataset, typically using vectorized representations and cosine similarity or other distance metrics to find the most similar sample videos.

Let $\mathbf{v}_{\text{target}}$ represent the embedding vector of the target video description, and $\mathbf{v}_i$ represent the embedding vector of the $i$-th sample video in the dataset. The semantic similarity is computed as:

\begin{equation}
\text{sim}(\mathbf{v}_{\text{target}}, \mathbf{v}_i) = \frac{\mathbf{v}_{\text{target}} \cdot \mathbf{v}_i}{\|\mathbf{v}_{\text{target}}\| \|\mathbf{v}_i\|}
\end{equation}

The category $C$ of the target video is determined by:

\begin{equation}
C = \arg\max_{c \in \mathcal{C}} \sum_{i \in I_c} \text{sim}(\mathbf{v}_{\text{target}}, \mathbf{v}_i)
\end{equation}

where $\mathcal{C}$ is the set of all categories and $I_c$ is the set of sample indices belonging to category $c$.

Once the video category is determined, the comment generation module begins to function. The first step of this module is to screen out the most style-matched comments from the video comment dataset of the corresponding category. To achieve this screening, the system employs a multi-layer voting mechanism. Specifically, first, 100 comments are extracted from the comment library and divided into 10 groups, with 10 comments per group.

Let $\mathcal{C}_{\text{pool}}$ be the initial comment pool with $|\mathcal{C}_{\text{pool}}| = 100$. The comments are partitioned into $G = 10$ groups:

\begin{equation}
\mathcal{C}_{\text{pool}} = \bigcup_{g=1}^{G} \mathcal{G}_g, \quad |\mathcal{G}_g| = 10, \quad \mathcal{G}_i \cap \mathcal{G}_j = \emptyset \text{ for } i \neq j
\end{equation}

Then, the system evaluates the style matching degree of comments in each group and selects the comment that best matches the target video style from each group. For each group $\mathcal{G}_g$, the best comment is:

\begin{equation}
c_g^* = \arg\max_{c \in \mathcal{G}_g} S(c, v)
\end{equation}

where $S(c, v)$ is the style matching score. Next, style matching degree evaluation is conducted again among these 10 candidate comments $\{c_1^*, c_2^*, \ldots, c_{10}^*\}$, and finally one most representative comment is selected as the "style template":

\begin{equation}
c^* = \arg\max_{c \in \{c_1^*, \ldots, c_{10}^*\}} S(c, v)
\end{equation}

The style matching score $S(c_j, v)$ for comment $c_j$ with respect to video $v$ can be formulated as:

\begin{equation}
S(c_j, v) = \alpha \cdot S_{\text{struct}}(c_j, v) + \beta \cdot S_{\text{tone}}(c_j, v) + \gamma \cdot S_{\text{length}}(c_j, v)
\end{equation}

where $S_{\text{struct}}$, $S_{\text{tone}}$, and $S_{\text{length}}$ represent structural, tonal, and length similarity scores respectively, and $\alpha$, $\beta$, $\gamma$ are weighting coefficients satisfying $\alpha + \beta + \gamma = 1$.

It is worth noting that this "style template" is not directly used for publication but serves as a reference structure for generating original comments. The system inputs this comment along with the target video's content description to a large language model (such as GPT-4o) and explicitly instructs the model to generate based on the comment's structure rather than its semantic content. The generation process can be formalized as:

\begin{equation}
c_{\text{generated}} = \text{LLM}(D_v, c^*, \mathcal{I})
\end{equation}

where $D_v$ is the video description, $c^*$ is the style template, and $\mathcal{I}$ represents the instruction to maintain structural style while creating original content. This means that the generated comment will maintain consistency with the template in terms of sentence structure, tone, and rhythm, but the specific expression will be entirely based on the target video's content, thus achieving true originality.

This generation method offers multiple advantages. First, it ensures consistent review style, making the generated results appear natural and relevant within the context. Second, by introducing multi-layered filtering and semantic matching mechanisms, the system effectively avoids issues such as style deviation or semantic inconsistency, ensuring strong robustness. Finally, leveraging the powerful generation capabilities of large-scale language models, the system can maintain style while creating novel and engaging review content, improving user experience and the quality of platform interaction.

Ultimately, through the collaborative work of multiple stages including video content understanding, semantic matching, style screening, and language generation, LOLGORITHM's comment generation module achieves automated transformation from video to high-quality comments. Its design philosophy emphasizes the balance between style consistency and content originality, reflecting high emphasis on user experience and content quality, and is an important manifestation of the system's intelligent level.
\section{Evaluation}

We implemented LOLGORITHM with around 1{,}200 lines of Python code.
The MLLM we used for the agents is GPT-4o. 
It is worth noting that in this paper, we do not focus on evaluating the performance of different MLLMs.
Therefore, we just evaluate the performance of LOLGORITHM powered by GPT-4o.
Any MLLMs with similar general capability as GPT-4o should allow LOLGORITHM to perform well.

 Our evaluation aims to answer the following two research questions:
\begin{itemize}[leftmargin=*]
    \item \textbf{RQ1:} Can LOLGORITHM generate comments similar to high-quality human-written comments?
    \item \textbf{RQ2:} How does LOLGORITHM perform compared to existing approaches in terms of human preferences?
\end{itemize}

\subsection{Evaluation Setup}

\subsubsection{Baseline Techniques}

We compare LOLGORITHM against three representative baselines that span different approaches to comment generation:

\begin{itemize}[leftmargin=*]
  \item \textbf{V2Xum-LLM}~\cite{Hua2024V2XumLLMCV} – A multimodal summarization model that focuses on extracting key information from video content to generate comments.
  \item \textbf{LiveChat}~\cite{Gao2023LiveChatAL} – A real-time interactive commentary system designed for live streaming scenarios.
  \item \textbf{GPT-4o (Direct)} – A baseline using the prompt structure of LOLGORITHM without the multi-agent collaborative framework, serving as an ablation baseline to demonstrate the effectiveness of the multi-agent framework.
\end{itemize}

These baselines represent different paradigms in comment generation: content summarization, real-time interaction, and direct large language model application.

\subsubsection{Benchmark Videos}

We constructed a dataset of 40 short videos crawled from Douyin (Chinese) and YouTube (English), with 20 videos from each platform.
These short videos are different from those used in the tool's knowledge base.
It is worth noting that we purposefully included several videos that do not fall into the categories of the tool's video classification to increase the challenge and simulate real-world scenarios, thereby evaluating how the tool performs on previously unseen types of videos.
Table~\ref{tab:videos} shows the detailed information of the videos used for benchmarking.
In addition to these videos, we also used the dataset construction module to collect five highly-rated comments for each video as reference standards for answering RQ1, and created a benchmark dataset for scoring in the automatic scoring framework.

\begin{table}[t]
\centering
\caption{Details of the videos used for benchmarking. (TS: Talk Show, etc.; HC: Humorous Commentary; FA: Funny Animal; DLJ: Daily Life Jokes; CS: Comedy Skits.}
\label{tab:videos}
\begin{tabular}{lllllll}
\hline
Platform & TS & HC & FA & DLJ & CS & Other \\ \hline
YouTube  & 3  & 3  & 3  & 4   & 3  & 4     \\ \hline
Douyin   & 3  & 4  & 2  & 4   & 3  & 4     \\ \hline
\end{tabular}
\end{table}

\subsubsection{Evaluation Metrics}

To comprehensively assess the performance of LOLGORITHM, we adopt a dual evaluation strategy combining automatic scoring and human preference analysis. 
This approach ensures our system is benchmarked both computationally and socially, capturing both objective metrics and subjective user engagement potential.
We employ two complementary evaluation strategies:

\paragraph{Automatic Scoring Framework}
We design an automatic scoring system that evaluates generated comments along three dimensions: originality, relevance, and style conformity. Each dimension has a maximum score of 10 points, and the final score is the average of the three dimensions:
$$S_{\text{total}} = \frac{S_{\text{originality}} + S_{\text{relevance}} + S_{\text{style}}}{3}$$
subject to $S_{\text{total}} \in [0,10]$ and $S_{\text{originality}}, S_{\text{relevance}}, S_{\text{style}} \in [0,10]$.

\textbf{Originality ($S_{\text{originality}}$, max 10):} 
For originality evaluation, we compare the similarity between each agent-generated comment and all comments in both the benchmark dataset and the original dataset used for comment generation, as well as with the content of the target video used for generating comments. Lower similarity scores result in higher originality scores, thereby ensuring the originality of generated comments. Specifically, if $c_i$ denotes a generated comment, $\mathcal{D}_{\text{bench}}$ represents all comments in the benchmark dataset, $\mathcal{D}_{\text{train}}$ represents all comments in the training dataset, and $v_i$ represents the target video content, we compute:
$$\text{sim}_{\text{max}} = \max\left(\max_{d \in \mathcal{D}_{\text{bench}} \cup \mathcal{D}_{\text{train}}} \text{sim}(c_i, d), \text{sim}(c_i, v_i)\right)$$
$$S_{\text{originality}} = 10 \cdot (1 - \text{sim}_{\text{max}})$$

\textbf{Relevance ($S_{\text{relevance}}$, max 10):} 
Relevance measures whether the generated comment is completely detached from the video topic or represents an unusable result due to hallucinations from the large language model. We first compute the similarity between existing comments in the benchmark dataset and their corresponding videos to obtain a baseline relevance value for human-written comments. Then, we calculate the similarity between generated comments and their corresponding videos and compare this value with the baseline. Scores are higher when the value is closer to the baseline, and lower when it deviates (either above or below) from the baseline. Formally, let $\text{sim}_{\text{baseline}}$ be the average similarity between human comments and videos in the benchmark:
$$\text{sim}_{\text{baseline}} = \frac{1}{|\mathcal{D}_{\text{bench}}|} \sum_{(c_j, v_j) \in \mathcal{D}_{\text{bench}}} \text{sim}(c_j, v_j)$$
For a generated comment $c_i$ and its corresponding video $v_i$:
$$S_{\text{relevance}} = 10 \cdot \exp\left(-\frac{|\text{sim}(c_i, v_i) - \text{sim}_{\text{baseline}}|^2}{2\sigma^2}\right)$$
where $\sigma$ is a scaling parameter that controls the sensitivity to deviation from the baseline.

\textbf{Style Conformity ($S_{\text{style}}$, max 10):} 
In the final dimension of style conformity, we incorporate word count judgment. We calculated the normal distribution of comment lengths in the benchmark dataset, separately for Chinese and English, and found that most human-written English comments fall within 63--72 words, while Chinese comments fall within 25--35 characters. Therefore, points are deducted if the generated comment's length falls outside these ranges. Additionally, we incorporate sentiment analysis using distilbert-base-uncased-finetuned-sst-2-english\cite{Sanh2019DistilBERTAD} for English content and BERT Base Chinese Finetuned Sentiment\cite{Devlin2019BERTPO} for Chinese content to analyze the video content, followed by sentiment analysis of the generated comments. Points are awarded if the two analysis results are consistent. The style score is computed as:
$$S_{\text{style}} = S_{\text{length}} + S_{\text{sentiment}}$$
where:
$$S_{\text{length}} = \begin{cases} 
5 & \text{if } L_{\min} \leq |c_i| \leq L_{\max} \\
5 \cdot \exp\left(-\frac{(|c_i| - L_{\text{nearest}})^2}{2\sigma_L^2}\right) & \text{otherwise}
\end{cases}$$
and $L_{\min}$, $L_{\max}$ are the length bounds (63--72 words for English, 25--35 characters for Chinese), $|c_i|$ is the length of comment $c_i$, and $L_{\text{nearest}}$ is the nearest bound.
$$S_{\text{sentiment}} = \begin{cases} 
5 & \text{if } \text{sentiment}(c_i) = \text{sentiment}(v_i) \\
0 & \text{otherwise}
\end{cases}$$

\paragraph{Human Preference Analysis}
We conducted a manual evaluation using a multimodal anonymous questionnaire. Each question consisted of a target video and four generated comments: one from each of the three baseline models and one from LOLGORITHM. Respondents watched the video and then selected the comment they were most likely to "like" or engage with, simulating real-world platform interaction. Two questionnaires, one in Chinese and one in English, were administered to short video enthusiasts aged between 20 and 50. The English questionnaire was completed by native English speakers, and the Chinese questionnaire was completed by native Chinese speakers.

\subsection{RQ1: Can LOLGORITHM generate comments similar to high-quality human-written comments?}

To answer this research question, we conducted quantitative analysis of LOLGORITHM and baseline methods through our automatic evaluation framework. Table~\ref{tab:douyin_results} and Table~\ref{tab:youtube_results} present detailed scoring results on the Douyin and YouTube datasets, respectively.

As shown in Table~\ref{tab:douyin_results}, on the Douyin dataset, LOLGORITHM achieved the highest total score of 5.55, significantly outperforming all baseline methods. Specifically, LOLGORITHM excelled in the relevance dimension with a score of 8.48, surpassing other methods (V2Xum-LLM: 8.24, LiveChat: 7.49, GPT-4o: 7.5), demonstrating its superior ability to maintain appropriate topical connection with video content while balancing the social interaction nature of short-video comments. In terms of style conformity, LOLGORITHM scored 6.07, substantially outperforming V2Xum-LLM (1.1) and GPT-4o (1.0), and exceeding LiveChat (5.44), indicating that its generated comments closely match the linguistic patterns, length requirements (25--35 characters for Chinese), and sentiment alignment expected on the Douyin platform. For originality, LOLGORITHM achieved 2.11, demonstrating moderate novelty while maintaining platform-appropriate style. The balanced performance across all three dimensions reflects LOLGORITHM's comprehensive approach to comment generation that neither sacrifices relevance for creativity nor conformity for originality.

On the YouTube dataset (Table~\ref{tab:youtube_results}), LOLGORITHM achieved the best overall performance with a total score of 5.42, outperforming LiveChat (4.92), V2Xum-LLM (4.5), and GPT-4o (4.02). LOLGORITHM demonstrated strong relevance with a score of 7.67, significantly exceeding LiveChat (4.9) and GPT-4o (4.29), though slightly lower than V2Xum-LLM's 8.67. This indicates LOLGORITHM's ability to generate contextually appropriate comments that connect meaningfully with video content. In originality, LOLGORITHM achieved the highest score of 3.26, surpassing V2Xum-LLM (3.12), LiveChat (3.08), and GPT-4o (2.78), showcasing its ability to generate fresh and distinctive comments. The style conformity score of 5.32 was competitive, positioned between LiveChat's leading 6.77 and V2Xum-LLM's lower 1.7, reflecting LOLGORITHM's strategy of balancing all three dimensions rather than optimizing for style alone. Notably, V2Xum-LLM's poor style conformity (1.7) despite high relevance (8.67) resulted in a lower total score (4.5), demonstrating the importance of balanced performance across all dimensions.

Overall, the automatic evaluation results demonstrate that LOLGORITHM has clear advantages in comment generation quality across both platforms, consistently achieving the highest total scores. Its multi-agent collaborative framework is particularly effective in maintaining strong relevance while generating original content and adapting to platform-specific style requirements. The results show that LOLGORITHM excels at balancing the three key dimensions—achieving high relevance scores (8.48 on Douyin, 7.67 on YouTube), the highest originality scores (2.11 on Douyin, 3.26 on YouTube), and strong style conformity (6.07 on Douyin, 5.32 on YouTube). These results validate that LOLGORITHM can generate comments similar to high-quality human-written comments, successfully balancing creativity, platform conformity, and topical connection.

\begin{table}[t]
\centering
\caption{Scoring Results on Douyin Dataset}
\label{tab:douyin_results}
\small
\begin{tabular}{p{2.5cm}cccc}
\hline
Model & Orig. & Rel. & Style & Total \\
\hline
V2Xum-LLM & 1.58 & 8.24 & 1.1 & 3.64 \\
LiveChat & 1.56 & 7.49 & 5.44 & 4.83 \\
\textbf{LOLGORITHM} & 2.11 & \textbf{8.48} & \textbf{6.07} & \textbf{5.55} \\
GPT-4o (Direct) & 0.7 & 7.5 & 1.0 & 3.06 \\
\hline
\end{tabular}
\end{table}

\begin{table}[t]
\centering
\caption{Scoring Results on YouTube Dataset}
\label{tab:youtube_results}
\small
\begin{tabular}{p{2.5cm}cccc}
\hline
Model & Orig. & Rel. & Style & Total \\
\hline
V2Xum-LLM & 3.12 & \textbf{8.67} & 1.7 & 4.5 \\
LiveChat & 3.08 & 4.9 & \textbf{6.77} & 4.92 \\
\textbf{LOLGORITHM} & \textbf{3.26} & 7.67 & 5.32 & \textbf{5.42} \\
GPT-4o (Direct) & 2.78 & 4.29 & 5.0 & 4.02 \\
\hline
\end{tabular}
\end{table}

\subsection{RQ2: How does LOLGORITHM perform compared to existing approaches in terms of human preferences?}

\begin{figure}[t]
    \centering
    \includegraphics[width=0.8\linewidth]{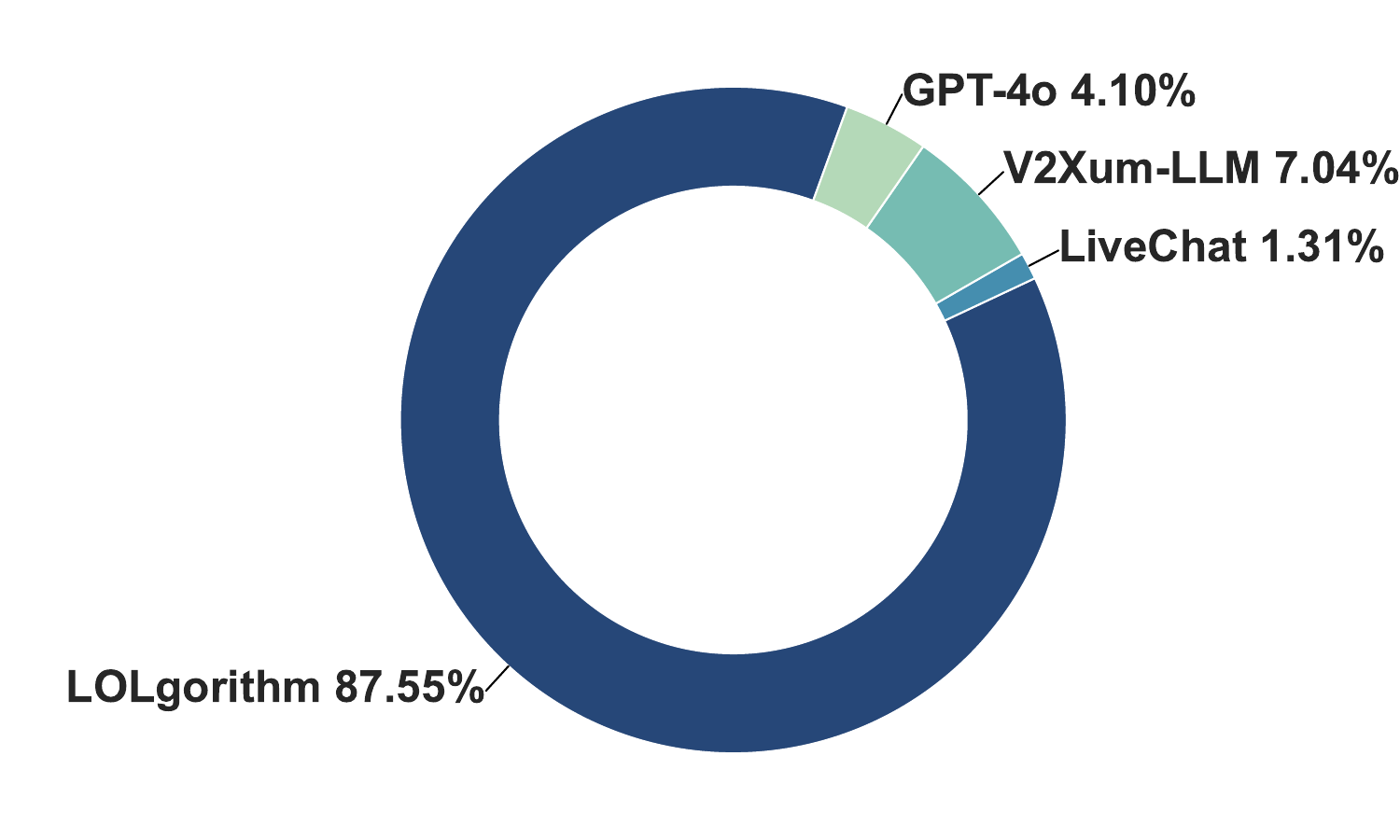}
    
    \medskip
    
    \includegraphics[width=0.8\linewidth]{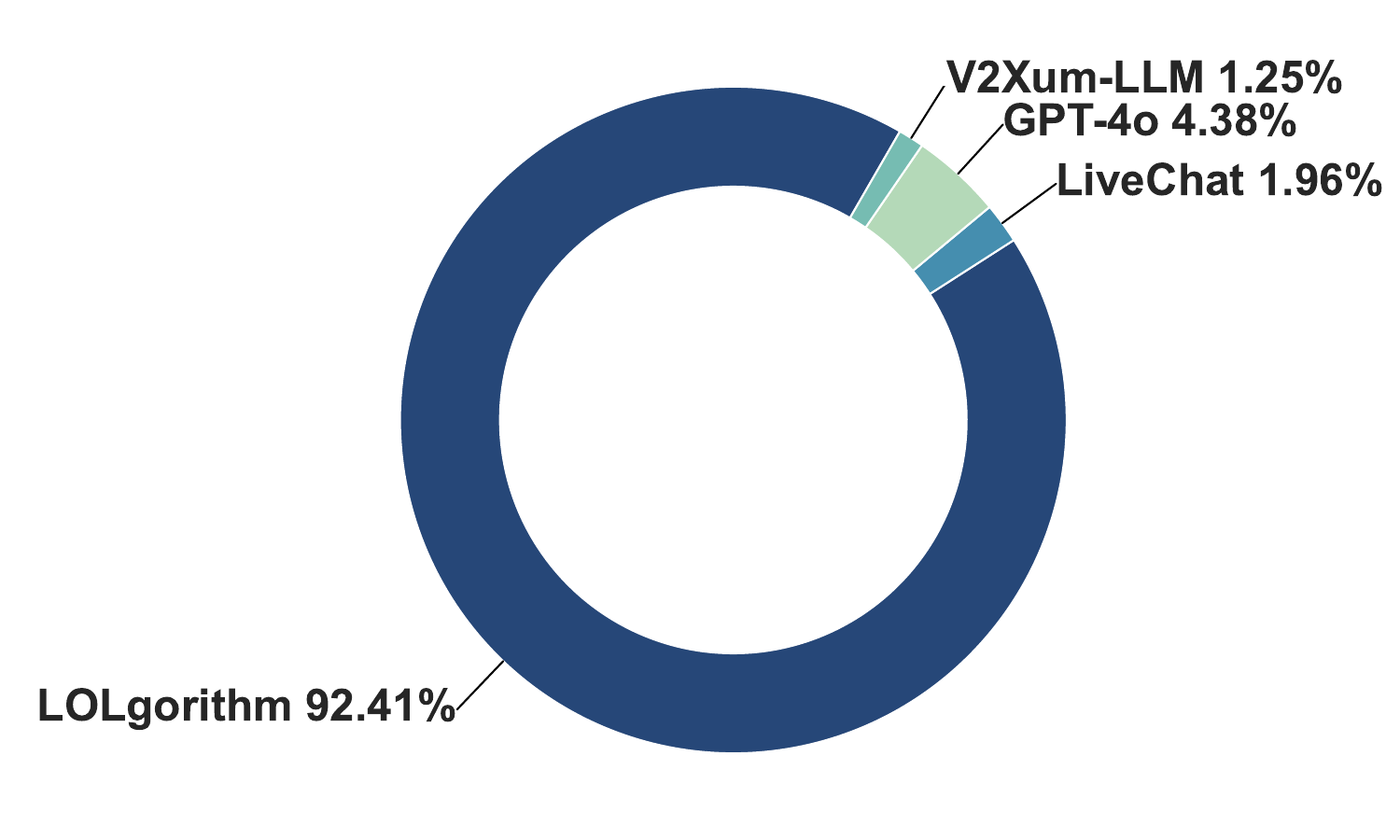}
    \caption{Human evaluation results for comment generation across platforms. 
             (a) YouTube results. (b) Douyin results.}
    \label{fig:human_eval_combined}
\end{figure}

To evaluate LOLGORITHM's performance in terms of real user preferences, we conducted a large-scale human evaluation experiment. Figure~\ref{fig:human_eval_combined} presents the human preference evaluation results on both Douyin and YouTube platforms.

The human evaluation results provide compelling evidence of our system's effectiveness. On the YouTube platform, with 51 valid responses collected across 20 video samples, LOLGORITHM was selected as the most preferred comment in 87.55\% of cases, significantly outperforming other baseline methods: LiveChat received only 1.31\% selection rate, V2Xum-LLM achieved 7.04\%, and GPT-4o (Direct) obtained 4.10\%. On the Douyin platform, the results were even more remarkable: through 56 valid responses across 20 video samples, LOLGORITHM achieved an overwhelming 92.41\% selection rate, while all other methods performed below 5\% (LiveChat: 1.96\%, GPT-4o: 4.38\%, V2Xum-LLM: 1.25\%).

These results clearly demonstrate that LOLGORITHM has significant advantages in generating user-preferred comments, with its multi-agent collaborative framework better capturing the comment culture and user expectations of different platforms.

\subsection{Case Study}

\begin{figure}[t]
    \centering
    \includegraphics[width=\linewidth]{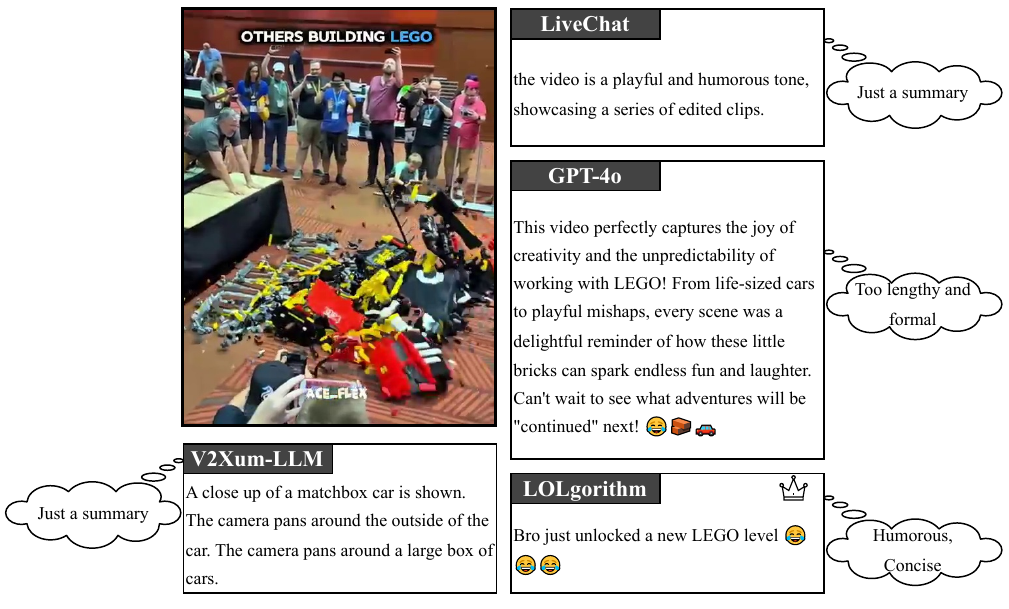}
    
    \medskip
    
    \includegraphics[width=\linewidth]{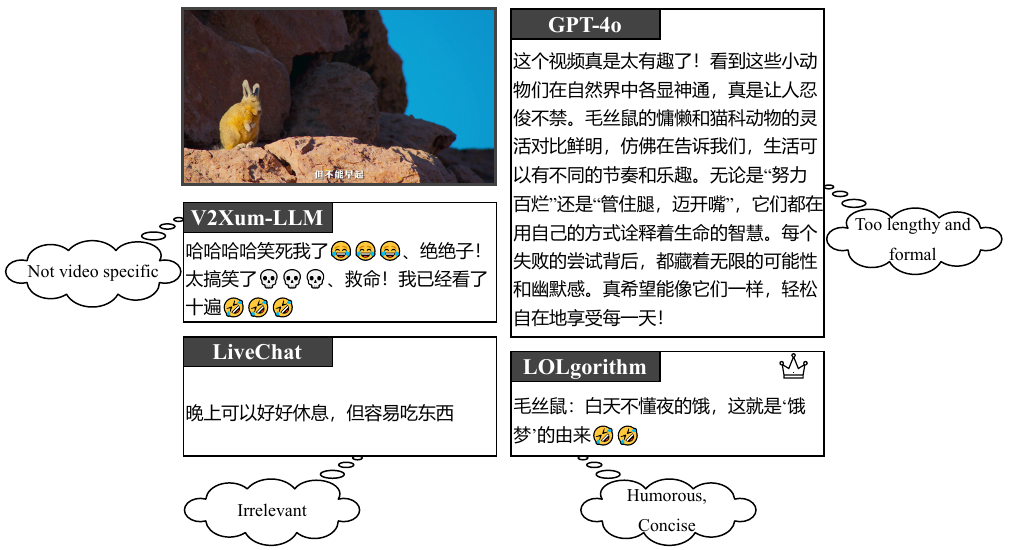}
    \caption{Example comments generated by different techniques. 
             Top: YouTube example. Bottom: Douyin example.}
    \label{fig:examples}
\end{figure}

Figure~\ref{fig:examples} further illustrates LOLGORITHM's advantages over baseline methods through specific YouTube and Douyin video cases.

In the YouTube case (Figure~\ref{fig:examples}Top), LOLGORITHM generated concise and humorous content that perfectly matches YouTube users' interactive style. In contrast, GPT-4o (Direct) was overly lengthy and formal, V2Xum-LLM only provided simple video descriptions, while LiveChat lacked specificity.

In the Douyin case (Figure~\ref{fig:examples}Bottom), LOLGORITHM cleverly used wordplay and internet slang, embodying the humorous style favored by Douyin users. GPT-4o (Direct), while rich in content, was overly formal and lengthy, not fitting Douyin's culture of concise and entertaining comments. V2Xum-LLM, though containing emojis, lacked specific relevance to the video content, while LiveChat was completely irrelevant.

These cases clearly demonstrate LOLGORITHM's superior ability to understand platform-specific comment cultures and generate comments of appropriate length and style, explaining why it achieved such high selection rates in human preference evaluation.

\section{Conclusion}




LOLGORITHM is a modular multi-agent system for generating platform-compliant, stylistically diverse short-video comments, achieving over 90\% user preference on Douyin and 87.55\% on YouTube. It introduces a novel framework combining video understanding with style-aware generation, a bilingual dataset of 1,000 annotated samples, and balanced performance across originality, relevance, and style. This work lays the groundwork for scalable, culturally adaptive comment generation, with future plans for multilingual expansion, personalization, real-time deployment, and open-sourcing to support broader research.
\bibliographystyle{ACM-Reference-Format}
\bibliography{references}










\end{document}